\documentclass[letterpaper, 10 pt, journal, twoside]{ieeetran}

\usepackage{graphics} 
\usepackage{graphics} 

\usepackage{times} 
\usepackage{amsmath} 
\usepackage{amssymb}  
\usepackage{dsfont}  
\usepackage{mathtools}
\usepackage{hyperref}
\usepackage[normalem]{ulem}
\useunder{\uline}{\ul}{}
\usepackage{multirow}
\usepackage{hyperref}
\usepackage{siunitx}

\usepackage{graphicx} 
\usepackage{subcaption}

\usepackage{pifont}

\usepackage[dvipsnames]{xcolor}
\usepackage{xspace}

\newcommand{\myparagraph}[1]{\vspace{3pt}\noindent\textbf{#1}}

\DeclareMathOperator*{\argmax}{arg\,max}

\IEEEoverridecommandlockouts           

\title{On the Challenges of Open World Recognition\\under Shifting Visual Domains} 

\author{Dario Fontanel$^{1}$, Fabio Cermelli$^{1,2}$, Massimiliano Mancini$^{3}$, 
and Barbara Caputo$^{1,2}$
\thanks{Manuscript received: October 15, 2020; Accepted December 12, 2020.}
\thanks{This paper was recommended for publication by Editor Cesar Cadena Lerma upon evaluation of the Associate Editor and Reviewers' comments.}
\thanks{This work was partially supported by Inxpect S.p.A (D.F.), the ERC grant 637076 - RoboExNovo (F.C., B.C.) and the German Research Foundation through the Cluster of Excellence “Machine Learning – New Perspectives for Science”, EXC 2064/1, project number 390727645 (M.M.) } 
\thanks{$^{1}$D. Fontanel, F. Cermelli and B. Caputo are with Politecnico di Torino, Turin, Italy. {\tt\small \{dario.fontanel, fabio.cermelli, barbara.caputo\}@polito.it}}%
\thanks{$^{2}$F. Cermelli, and B. Caputo are with Italian Institute of Technology, Genoa, Italy.}
\thanks{$^{3}$M. Mancini is with University of T\"ubingen, T\"ubingen, Germany. {\tt\small massimiliano.mancini@uni-tuebingen.de}}
}

\begin{document}

\fbox{\parbox{.99\textwidth}{
{\LARGE
\textbf{Disclaimer:}} \\

{\large
This work has been accepted for publication at IEEE Robotics and Automation Letters and IEEE International Conference on Robotics and Automation (ICRA) 2021. \\

link: https://www.ieee-icra.org/ \\

If you want to cite this article, please use:\\
\textit{
@article\{fontanel2021challenges, \\
  author=\{Fontanel, Dario and Cermelli, Fabio and Mancini, Massimiliano and Caputo, Barbara\}, \\
  journal=\{IEEE Robotics and Automation Letters\}, \\
  title=\{On the Challenges of Open World Recognition Under Shifting Visual Domains\}, \\
  year=\{2021\}, \\
  volume=\{6\}, \\
  number=\{2\}, \\
  pages=\{604-611\}, \\
  doi=\{10.1109/LRA.2020.3047777\} \\
 \}\\
} 

{\large 
Copyright:\\
© 2021 IEEE. Personal use of this material is permitted. Permission from IEEE must be obtained for all other uses, in any current or future media, including reprinting/republishing this material for advertising or promotional purposes, creating new collective works, for resale or redistribution to servers or lists, or reuse of any copyrighted component
of this work in other works.
}
}}}

\maketitle

\begin{abstract}
Robotic visual systems operating in the wild must act in unconstrained scenarios, under different environmental conditions while facing a variety of semantic concepts, including unknown ones. To this end, recent works tried to empower visual object recognition methods with the capability to i) detect unseen concepts and ii) extended their knowledge over time, as images of new semantic classes arrive. This setting, called Open World Recognition (OWR), has the goal to produce systems capable of breaking the semantic limits present in the initial training set. However, this training set imposes to the system not only its own semantic limits, but also environmental ones, due to its bias toward certain acquisition conditions that do not necessarily reflect the high variability of the real-world. This discrepancy between training and test distribution is called domain-shift. This work investigates whether OWR algorithms are effective under domain-shift, presenting the first benchmark setup for assessing fairly the performances of OWR algorithms, with and without domain-shift. We then use this benchmark to conduct analyses in various scenarios, showing how existing OWR algorithms indeed suffer a severe performance degradation when train and test distributions differ. Our analysis shows that this degradation is only slightly mitigated by coupling OWR with domain generalization techniques, indicating that the mere plug-and-play of existing algorithms is not enough to recognize new and unknown categories in unseen domains. Our results clearly point toward open issues and future research directions, that need to be investigated for building robot visual systems able to function reliably under these challenging yet very real conditions.
\end{abstract}

\begin{IEEEkeywords} 
Deep Learning for Visual Perception, Visual Learning, Recognition
\end{IEEEkeywords}

\section{Introduction} 
\label{sec:intro}

\IEEEPARstart{G}{iven} an image, recognizing the presence of an object and its semantic category is a fundamental capability 
for any robotic visual system. Indeed, knowledge about the category is helpful in many 
tasks, such as object manipulation \cite{kragic2005vision}, handling \cite{saxena2008robotic}, and kitting \cite{pavlichenko2018kittingbot}.  A primary issue of standard object recognition algorithms is the closed world assumption (CWA), meaning that the set of categories available during training are assumed to be the only one the robot will ever encounter when deployed. This assumption is unrealistic for robots operating in unconstrained scenarios, due to the infinite number of semantics present in the world. 

The urge to break the CWA has lead researchers to consider the \textit{open world} recognition (OWR) problem \cite{bendale2015towards}. In OWR, an algorithm is asked to both detect unseen semantic concepts as well as to learn new semantic categories over time. Various solutions for OWR have been developed \cite{bendale2015towards,de2016online,mancini2019knowledge,fontanel2020boosting}, based on shallow \cite{bendale2015towards,de2016online} and deep \cite{mancini2019knowledge,fontanel2020boosting} classification models. 

\begin{figure}[tb]
  \centering
  \includegraphics[width=\columnwidth]{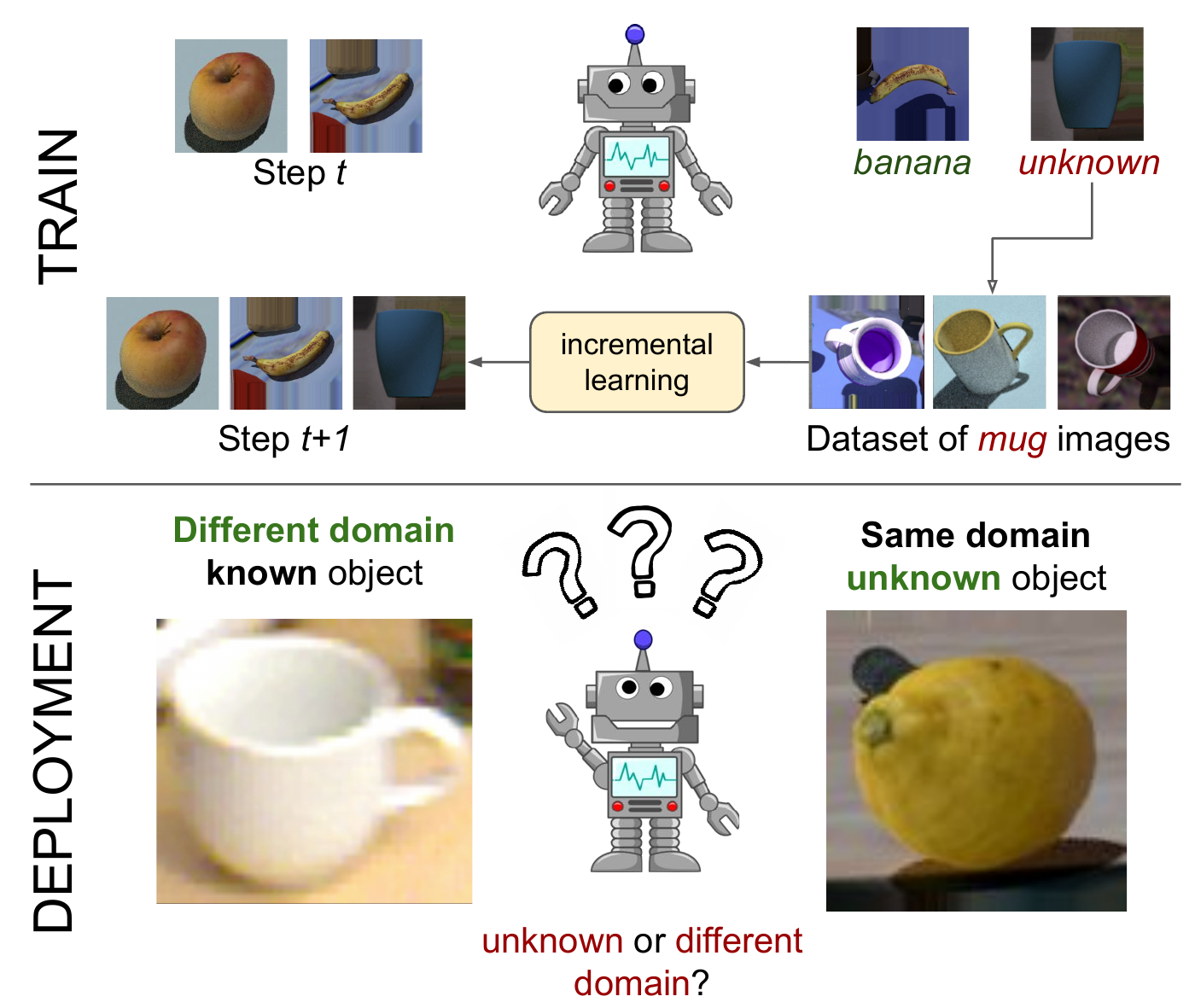}
  \vspace{-10pt}
   \caption{Our considered problem. In OWR a robot is asked to incrementally learn new concepts over time while detecting images containing unseen concepts. Our question is: does the effectiveness of the visual system hold when acting in different visual domains and environments?}
  \vspace{-10pt}
  \label{fig:teaser}
 \end{figure}

Despite their effectiveness, these algorithms always consider the training and test images to share the same acquisition conditions. As for CWA, this assumption, which we name closed domain (CDA), is sound only for robots operating in highly constrained settings, such as industrial robots. However, CDA is unrealistic for e.g., mobile robots working in the wild, and we need their visual systems to be able of dealing with the various input distributions (a.k.a, domains) that can arise from, e.g., different environments, illumination, and acquisition conditions. As an example, a visual system for security robots patrolling public spaces, trained on purely day-time images, might struggle to generalize to night-time ones due to abrupt differences between the two. 
This difference between training and test distributions is called domain-shift \cite{csurka2017comprehensive}. Multiple works addressed this problem in robot vision, under the framework of domain adaptation (DA) \cite{angeletti2018adaptive,loghmani2020unsupervised}. In the standard domain adaptation scenario, we have labeled data for one training (\textit{source}) domain, and unlabeled data for one test (\textit{target}) domain, and the goal is to use these data to model the discrepancy among source and target distributions. Since the presence of target data beforehand is a strong requirement, recent works tried to perform adaptation without target data during the initial training stage. In particular, they either exploit the stream of incoming target data \cite{mancini2018kitting,wulfmeier2018incremental}, a domain description \cite{yang2016multivariate,mancini2019adagraph}, and generalization strategies on either single \cite{volpi2019addressing,mancini2017learning} or multiple \cite{mancini2018robust} source domains. Despite these efforts, most of the DA algorithms focused on the case where training and test categories are shared, i.e., under the CWA. 

Since our final goal is to break CWA and CDA simultaneously (see Fig. \ref{fig:teaser}), a crucial open question remains: can OWR algorithms work under domain-shift? In this work, we try to answer this question, benchmarking OWR algorithms under changes between training and test distributions.

To fulfill this goal, we test three OWR algorithms, namely Nearest-Non Outlier \cite{bendale2015towards,de2016online}, DeepNNO \cite{mancini2019knowledge} and Boosting Deep OWR by Clustering (B-DOC) using the well-known RGB-D Object dataset (ROD) \cite{lai2011large} and two datasets sharing the same semantic categories but different acquisition conditions, namely synthetic ROD (synROD) \cite{loghmani2020unsupervised} and Autonomous Robot Indoor Dataset (ARID) \cite{loghmani2018recognizing}.
Training the models on either synROD or ROD and testing on the other datasets, we show how OWR algorithms indeed suffer from severe performance degradation when tested on domains different from the training one, with drops ranging from 10\% to almost 45\% in OWR harmonic mean. Interestingly, despite the highest performance on in-domain test, end-to-end trained deep OWR algorithms show to suffer the domain-shift problem even more than non end-to-end counterparts.

We then couple NNO, DeepNNO, and B-DOC with three single source domain generalization (DG) algorithms. Our experiments show that 
DG algorithms mitigate but do not resolve this problem, underlying how the objective of solving CWA and CDA together is far from being solved.

\myparagraph{Contributions.} To summarize:
\begin{itemize}
    \item We perform the first benchmark of OWR algorithms under domain-shift, showing how their performances heavily decrease when tested on different domains.
    \item We show how coupling OWR models with single-source DG ones can only reduce but not eliminate this problem.
    \item We propose a validation procedure and we release our benchmark for allowing easy and fair future research.
\end{itemize}



\section{Related Works} 
\label{sec:relateds}
\subsection{Open World Recognition}
The seminal work of Bendale et al. \cite{bendale2015towards} defined the OWR problem, trying to overcome the clear limitations of the CWA. In \cite{bendale2015towards}, the authors also proposed the first OWR algorithm, NNO, extending the popular Nearest Class Mean (NCM) \cite{mensink2012metric} with a rejection option for detecting unseen concepts. In \cite{de2016online} an extension of NNO is proposed for dynamic update of the classifier with online data streams. 

Both \cite{bendale2015towards} and \cite{de2016online} developed their OWR algorithms on top of shallow features. In our previous works \cite{mancini2019knowledge, fontanel2020boosting} we tested the benefits of more powerful deep representations \cite{alexnet}, developing two deep learning-based approaches for OWR. In \cite{mancini2019knowledge}, we proposed the DeepNNO algorithm, an end-to-end trainable deep extension of the original NNO using Deep NCM \cite{guerriero2018deep} as classifier. 
In \cite{fontanel2020boosting}, we presented B-DOC, an algorithm improving DeepNNO by considering clustering objectives and the use of class-specific rejection thresholds. 

Despite the advances in the field, none of these works tested the robustness of the models under domain-shift. Preliminary experiments were performed only on \cite{mancini2019knowledge}, showing that using web images to learn novel classes was feasible but with a decrease in the overall performance of the model, due to the label noise and the domain-shift. In this work, we are the first to explicitly benchmark these algorithms under settings involving different training and test distributions. 

\subsection{Domain-Shift in Robot Vision}
For robotic systems working in the wild is of utmost importance to develop models robust to domain-shift.
With this goal, several efforts have been devoted in robotics to perform adaptation in the presence of target data \cite{wulfmeier2017addressing,fang2018multi,angeletti2018adaptive,james2019sim,jeong2020da-manipulation,zhang2019vr}. 
However,
since it is impossible to collect data for every possible target domain beforehand, researchers explored techniques to address the domain-shift problem without the presence of target data during training. One solution is using the stream of incoming target data, performing adaptation in an online fashion by e.g., updating domain-specific components \cite{mancini2018kitting} and/or through adversarial objectives \cite{wulfmeier2018incremental}. A drawback of these strategies is their slow adaptation time, a problem in scenarios where fast adaptation is required (e.g., sudden illumination changes).  

A way to produce systems robust to (possibly) any target domain without using any target data (neither during training nor during employment), is resorting to domain generalization techniques \cite{mancini2018robust}. In this scenario, standard methods focused on the multi-source settings 
\cite{mancini2018robust}, while less research has been devoted to the case when only a single source domain is present. 
In the latter, since we cannot explicitly disentangle domain-specific and semantic-specific information, one solution is to build structurally more robust classifiers by means of part-based models \cite{urvsivc2016part,mancini2017learning}, multiple visual cues \cite{pal2019deduce}, regularization strategies \cite{huang2020selfchallenging} and self-supervised learning \cite{carlucci2019domain}. Another option is to simulate the presence of multiple source domains through adversarial techniques \cite{volpi2018adversarial} or data augmentation \cite{volpi2019addressing}. In the latter case, data augmentation can simulate increasingly harder new domains \cite{volpi2019addressing} or fictitious multiple sources \cite{qiao2020learning}.

In this work, we focus on DG models due to their readily applicability to various target domains without requiring any target data. Despite some efforts in testing domain adaptation models under open-set \cite{panareda2017open,bucci2020rotation} and zero-shot \cite{mancini2020dgzsl} scenarios, we are the first to explicitly study the domain-shift problem in OWR, testing the effectiveness of coupling OWR and DG algorithms for it. 

\section{Benchmarking OWR Algorithms under Domain-Shift} 
\label{sec:benchmark}
\subsection{Problem formulation}
Let us formalize the OWR problem. Suppose we have an initial training set $\mathcal{T}_0={(x_i,y_i)}_{i=1^{N_0}}$, with $x_i$ being an image in the image space $\mathcal{X}$, $y_i$ being a class label in the set $\mathcal{Y}_0$, and $N_0$ the number of samples. Sequentially, at learning step $T$ we will receive another training set $\mathcal{T}_T$ containing a novel set of classes, i.e., $\mathcal{Y}_T \bigcap \mathcal{Y}_t=\emptyset\; \forall t \in [0,T-1]$. Our goal is to learn a function  $f:X\rightarrow \mathcal{K}_t \cup \text{u}$ mapping an image $x$ into either one of the semantic classes learned until step $T$ (i.e., $\mathcal{K}^T=\bigcup_{t=0}^T\mathcal{Y}_t$) or the unknown class $\text{u}$. 

Note that, $f$ will be incrementally updated (as a new training set arrives) while still being asked to detect possibly unseen concepts. OWR algorithms differ in the way $f$ is defined and learned. Without loss of generality, we consider $f$ being built on three components: a feature extractor $\omega:\mathcal{X}\rightarrow\mathcal{Z}$ mapping images into a feature space $\mathcal{Z}$; a scoring function $\phi:\mathcal{Z}\rightarrow\Re^{|\mathcal{K}^T|}$ mapping features in $\mathcal{Z}$ to known class scores; and $\sigma:\Re^{|\mathcal{K}^T|}\rightarrow \mathcal{K}_t \cup \text{u}$, mapping the class scores to the final prediction. In the following we will describe how various OWR algorithms have defined (and eventually learned) $\omega$, $\phi$ and $\sigma$ (see Table \ref{tab:owr} for a summary).

\begin{table}[]
\vspace{+4pt}
\centering
\begin{tabular}{l|l|l|l}
 & NNO & DeepNNO & B-DOC \\ \hline
$\omega$ & fixed & updated & updated \\
$\phi$ & $\mathcal{N} (1-\frac{d(z, \mu_y)}{\tau})$ & $\exp\left(-\frac{1}{2}||z - \mu_y||\right)$ & $\frac{1}{\varphi} {||z - \mu_y||}^2$ \\
$\sigma$ & $\tau \le 0$ & $\phi \le \tau$ & $\phi > \tau$ \\
$\tau$ & fixed & updated & learned \\ \hline
\end{tabular}
\vspace{-2pt}
\caption{Difference among OWR algorithms. Each method learns a classification function  $f$ in which $\omega$ is the feature extractor, $\phi$ is the scoring function, $\sigma$ is the final prediction function, $\mathcal{N}$ is a normalization factor, $\varphi$ where is the standard deviation of the features in $z=\omega(x)$ and $\tau$ is the method-specific threshold(s).
\vspace{-8pt}
}
\label{tab:owr}
\end{table}

\subsection{OWR algorithms}
\myparagraph{Nearest Non-Outlier (NNO).}
NNO \cite{bendale2015towards} is a non-parametric OWR approach. For a known class $y$ and a sample $x$, the scoring function $\phi$ takes the form:
\[
\phi_y^\text{NNO}(z)=\mathcal{N} (1-\frac{d(z, \mu_y)}{\tau}), 
\]
where $z=\omega(x)$, $d$ is a distance measure, $\mu_y$ is the class-specific centroid, $\tau$ is a rejection threshold and $\mathcal{N}$ is a normalization factor. The class specific centroid $\mu_y$ is computed with the Nearest-Class Mean (NCM) algorithm \cite{mensink2012metric}, while the threshold $\tau$ using a set of held-out validation samples. The final prediction is obtained by: 
\[
    \sigma(z) = 
  \begin{cases}
    \text{u} & \text{if}\; \phi_y^\text{NNO}(z) \leq 0\; \forall{y \in \mathcal{K}_t},\\
    \argmax_{y \in \mathcal{Y}_T} \phi_y^\text{NNO}(z) & \text{otherwise}.
  \end{cases}
\]
While in the general case, $\omega$ is a shallow feature extractor (e.g., SIFT \cite{lowe1999object}), in our benchmark we use a deep architecture pretrained on the base training set $\mathcal{T}_0$ with DeepNCM \cite{guerriero2018deep} as classifier and the online version of NNO presented in \cite{de2016online} to compute the class centers $\mu_y$ and the threshold $\tau$. 

\myparagraph{Deep Nearest Non-Outlier (DeepNNO).}
DeepNNO \cite{mancini2019knowledge} is a deep extension of NNO. In this case, $\omega$ is a deep architecture, end-to-end trained. The score function is:
\[
\phi_y^{\text{DNNO}}(z)=\exp\left(-\frac{1}{2}||z - \mu_y ||\right) .
\]
with the final prediction obtained as:
\[
    \sigma(z) = 
  \begin{cases}
    \text{u} & \text{if}\; \phi_y^\text{DNNO}(z) \leq \tau\; \forall{y \in \mathcal{K}_t},\\
    \argmax_{y \in \mathcal{Y}_T} \phi_y^\text{DNNO}(z) & \text{otherwise}.
  \end{cases}
\]
Both $\mu_y$ and $\tau$ are updated online, the first as the underline feature representation is learned, the second based on the confidence on the scores for both correctly and wrongly classified training samples. At step $t$, the feature extractor $\omega$ is trained by minimizing the following objective:
\begin{equation}
\label{eq:loss-dnno}
 \mathcal{L}=\frac{1}{|\mathcal{T}_t|}\sum_i\ell_{BCE}(x_i,y_i)  + \lambda  \ell_{DS}(x,\omega^{t-1})
\end{equation}
where $\lambda$ is a trade-off hyperparameter, $\ell_{BCE}$ is the standard binary-cross entropy loss 
%
 and $\ell_{DS}$ the distillation loss defined in \cite{hinton2015distilling} on the feature space:
\begin{equation}
 \label{eq:DS}
 \ell_{{DS}}(x,\omega_{t-1}) =  ||\omega(x)- \omega^{t-1}(x)||
\end{equation}
with $\omega^{t-1}$ being the feature extractor after the previous learning step. This loss, together with a set of stored samples of old classes prevents forgetting of past knowledge.

\myparagraph{Boosting Deep OWR by Clustering (B-DOC).}
B-DOC \cite{fontanel2020boosting} revises the ideas of DeepNNO by imposing two clustering constraints on the feature space and learning class-specific rejection thresholds. In particular, the score function of B-DOC is directly defined on the distances from the features and the class centroid, i.e. it is defined as:
\begin{equation} \label{eq:score-bdoc} 
\phi_y^\text{BDOC}(z)=\frac{1}{\varphi} {||z - \mu_y||}^2,
\end{equation}
where $\varphi$ is the standard deviation of the features in $z=\omega(x)$. The prediction function is: 
\[
    \sigma(x) = 
  \begin{cases}
    \text{u} & \text{if}\, \phi_y^\text{BDOC}(z)> \tau_y,\; \forall{y \in \mathcal{Y}_t},\\
    \text{argmin}_y \phi_y^\text{BDOC}(z) &\text{otherwise}
  \end{cases}
\]
with $\tau_y$ being a class-specific rejection threshold learned using a maximal distance constraint in a reserved set of training samples with random augmentations. While the distillation loss of Eq.~\eqref{eq:DS} and the stored exemplars are used to prevent forgetting, the representation is enforced with two clustering objectives. The first is the cross-entropy loss on the softmaxed negative scores computed in Eq.~\eqref{eq:score-bdoc}, for global clustering. The second is the soft-nearest neighbor loss \cite{frosst2019analyzing} for local clustering:
\[
    \ell_{SNNL}(x, y,\mathcal{B})  = - \log\ \  
    \frac{ \displaystyle\sum_{\mathclap{x_j \in \mathcal{B}_y\setminus \{x\}}
                               }
                      e^{- \frac{1}{\varphi}||\omega(x) - \omega(x_j)||^2}}{
          \displaystyle\sum_{\mathclap{x_k \in \mathcal{B}\setminus \{x\}}}
             e^{- \frac{1}{\varphi}||\omega(x) - \omega(x_k)||^2}
          }
\]
with $\mathcal{B}$ being the current training batch, and $\mathcal{B}_y$ the set of samples in the training batch belonging to class $y$. The final loss is:
\[
 \mathcal{L}=\frac{1}{|\mathcal{T}_t|}\sum_i\ell_{CE}(x_i,y_i)+\gamma \ell_{SNNL}(x_i,y_i,\mathcal{B})  + \lambda  \ell_{DS}(x,\omega^{t-1})
\]
with $\gamma$ and $\lambda$ being trade-off hyperparameters.

\subsection{Single-source DG algorithms}
In this section, we describe the single source domain generalization algorithms we use in our benchmark. 

\myparagraph{Data augmentation with transformation sets. (RSDA)}
The first common approach for addressing single source domain generalization is through data augmentation techniques, either adversarial \cite{volpi2018adversarial,qiao2020learning} or transformation based \cite{volpi2019addressing}. As representative of this category, we choose the data augmentation based approach of \cite{volpi2019addressing}. In particular, given a training batch $\mathcal{B}=\{(x_i,y_i)\}_{i=1}^n$, the model is trained applying the semantic objectives on a transformed version of the batch $\mathcal{\hat{B}}=\{(\alpha{x}_i,y_i)\}_{i=1}^n$ where $\alpha$ is a randomly sampled transformation from a set $\mathcal{A}$. Given a set of simple transformations $A$ (e.g., blurring, mirroring) the set $\mathcal{A}$ is populated by composed transformations using the elements in $A$. In particular, an evolutionary-based search selects the combinations of $A$ leading to the worst performances for the current model and adds it to $\mathcal{A}$. The hyperparameters of the model are the set of basic transformations $A$, their possible values and the frequency at which $\mathcal{A}$ is updated. We fix the set $A$ to the following transformations: hue, contrast, brightness, saturation, random crop and mirroring.

\myparagraph{Self-supervised learning with relative rotations. (RR)}
Another popular strategy for achieving good domain generalization performances is through self-supervised learning \cite{carlucci2019domain}. In particular, the presence of a self-supervised auxiliary task makes the model focus on discriminative invariances and regularities helping generalization to new domains \cite{carlucci2019domain}. For the task to be helpful, it must require the model to reason on the actual content of the image, rather than its specific style and appearance. Among possible tasks, effective ones are solving jigsaw puzzles \cite{noroozi2016unsupervised,carlucci2019domain} and predicting rotations \cite{gidaris2018unsupervised,bucci2020rotation}. Here we take the task of relative rotations \cite{bucci2020rotation}. In particular, given a batch $\mathcal{B}$, we build a new batch $\mathcal{\hat{B}}$ as $\mathcal{\hat{B}}=\{(x_i,y_i, \text{rot}_{\theta_i}(x_i),\theta_i)\}_{i=1}^n$ where $\text{rot}$ is a rotation transformation applied with angle $\theta_i$ to the original image $x_i$. Since $\theta_i$ is sampled from a discrete set $\Theta$ (i.e., \ang{0}, \ang{90}, \ang{180} and \ang{270}), the auxiliary task is classifying which $\theta_i$ has been applied to $x_i$. To perform this, we instantiate a new network branch $\rho$ mapping features extracted from the original image and its rotated counterpart to the correct rotation angle, i.e., $\rho:\mathcal{Z}\times\mathcal{Z}\rightarrow\Theta$. A standard cross-entropy loss is used on top of the rotation predictions and it is used to update both $\rho$ and $\omega$. In the full objective of an OWR algorithm (e.g., Eq.\eqref{eq:loss-dnno}), this auxiliary loss is added scaled by a trade-off parameter $\xi$. Note that we apply the semantic objectives also to the rotated samples and that we perform random data augmentations to them, to increase the complexity of the auxiliary task. 

\myparagraph{Regularization through self-challenging. (SC)}
Finally, regularization strategies can improve generalization to unseen domains \cite{huang2020selfchallenging}. Here we test the self-challenging algorithm of \cite{huang2020selfchallenging}, where the model is asked to classify corrupted features, obtained by removing the elements that mostly contributed to a correct classification of the current sample. Formally, features are extracted from the original samples (i.e., $z=\omega(x)$), to compute a score $\phi_y(z)$ for the ground-truth class $y$. Then, the gradient of the score with respect to $z$ is computed (i.e., $g = \partial{\phi_y(z)}/\partial{z}$). Finally, a new set of features $\hat{z}$ is calculated by applying a mask $m$ on the original features $z$, i.e., $\hat{z}=m\circ z$ where $\circ$ is the Hadamard product and $m$ is a binary mask with 0s for every value $z_j$ whose gradient $g_j\geq q_p$, and 1 otherwise. The threshold $q_p$ is computed ensuring to preserve the top-p percentile of the activations per corrupted samples. The hyperparameters of the model are the corruption ratios sample- and batch-wise.

\subsection{Experimental setting}
We conduct the experiments on three datasets 
: RGB-D Object dataset (ROD)~\cite{lai2011large}, synthetic ROD (synROD)~\cite{loghmani2020unsupervised}, and Autonomous Robot Indoor Dataset (ARID)~\cite{loghmani2018recognizing}. The three datasets contain images of the same set of 51 daily-life objects but under very different acquisition conditions. 

\textbf{ROD} \cite{lai2011large}, is one of the most used dataset for object recognition in robotics. In ROD, different instances of the objects are captured while lying on a table in a quasi-ideal scenario, where there is no clutter, no illumination or background changes but only different camera angles. 

\textbf{synROD} \cite{loghmani2020unsupervised} is a synthetic version of ROD \cite{lai2011large}, {created by rendering public 3D models available on free catalogs. The authors rendered the scenes using a ray-tracing engine in Blender to simulate a realistic lighting.} This benchmark was proposed to test the ability of a model to deal with the domain-shift existing between synthetic and real images. 

In \textbf{ARID} \cite{loghmani2018recognizing} instead 
the objects are represented with several backgrounds, scales, views, lighting conditions, and different levels of occlusions. Thus, ARID is a more challenging dataset, originally developed to evaluate the robustness of deep recognition models in unconstrained environments. 

Regarding the evaluation procedure, we divide the semantic categories following the same split provided in \cite{fontanel2020boosting}, i.e., we use 26 classes as known and 25 as unknown, considering the first 11 classes as base classes and adding 5 classes at the time in each incremental step. In each experiment we used {all the images belonging to} the instances of the first train-test split defined in \cite{lai2011large}, with one instance per class in the test set and all the others in the training set. 
{
For synROD, we followed the split proposed in \cite{loghmani2020unsupervised}, adding the images of the 3 classes of ROD excluded from the benchmark\footnote{Images of the remaining 3 classes were provided by the authors of \cite{loghmani2020unsupervised}.}. 
Finally, we use ARID \cite{loghmani2018recognizing} 
only for test, being the most challenging and realistic scenario.
} 


\myparagraph{Metrics.} 
To asses the performances of OWR methods we use two different metrics. To evaluate the ability to learn new concepts, we use the accuracy over the set of known classes (i.e, closed world) averaged across all incremental steps. We report the results for this metric with and without the rejection option (i.e., the possibility to detect unknown concepts). To test the performances on the actual OWR scenario, 
we use the open world harmonic mean (OWR-H), i.e., the harmonic mean between the closed world with rejection and the open set accuracies averaged across all incremental steps, as proposed in \cite{fontanel2020boosting}.


\subsection{Validation protocol}
An open question in OWR is how to set the values of the hyperparameters of a method since i) only a subset of the semantic classes is available in each training step and ii) images of unknown categories are not available. 
Here, we propose a strategy exploiting only the base classes to find the best hyperparameters. 


In particular, we propose to split the classes available in the first training stage in two sets: known and unknown classes. We consider 10\% of the base classes (e.g., 2 out of 11 in our benchmark) as unknown while the rest as known classes. From the known class set, we use 50\% of them (e.g., 5 out of 9 in our benchmark) as classes for the first training step, 
while we use the second half (4 out of 9 in our case) to construct the incremental learning steps.
Note that with these three splits we have artificially created i) a set of base classes to start training the model, ii) a set of classes that will be incrementally learned and iii) a set of unseen concepts to evaluate the open set performances of the model. Finally, since the number of classes we will receive in each incremental step is unknown during deployment, we simulate this uncertainty by adding the set of incremental classes in multiple trials with different class cardinality. Specifically, we use multiple steps with a single class (e.g., 4 steps with 1 class in our benchmark), two steps with half of the classes (e.g., 2 steps of 2 classes), and a single step with all of them (e.g., 1 step with 4 classes). The split among known/unknown classes and among base/incremental is repeated multiple times, to ensure a better hyperparameter values estimation.

Once we obtain these base class splits, we perform hyperparameters validation in two steps. In the first place, the hyperparameters that contribute \textit{only} to acquire the knowledge of the network (i.e., closed world without rejection) are validated. This ensure that the method is capable of actually learning the novel concepts, without the risk of focusing on either retaining only the old knowledge or having low confidence on the predictions which might damage the open set performance lately. In this stage, two kind of hyperparameters are validated: the ones related to the network optimized (e.g., learning rate and weight decay) and the ones related to the weights of the semantic loss functions (e.g., cross-entropy, distillation and clustering). In the second step, we validate all hyperparameters related to the detection of samples containing unknown categories, using the OWR-H performances. For instance, we validate the negative weight used to update the rejection thresholds in DeepNNO \cite{mancini2019knowledge} and the learning rate used by B-DOC \cite{fontanel2020boosting} to learn the class-specific rejection thresholds.

We want to highlight that this whole procedure is agnostic to the underlying OWR model and the benchmark, using just the set of base classes to select the optimal hyperparameters. The only hyperparameters that we did not set using this protocol are the training epochs of base and incremental steps. For these, we select the epoch by looking at the training accuracy on the classes present in each learning step. 
In our case, we set 12 epochs for ROD and 70 for synROD for the base classes. For the incremental step we use a number of epoch proportional to the number of added classes for ROD, while we fix this value to 35 for synROD, being a more difficult scenario. 

\section{Results} 
\label{sec:experiments}
In this section, we show the results of our benchmark. We start by testing standard OWR algorithms under domain-shift (Section \ref{exp:owr-alone}), in both Syntethic-to-Real and Constrained-to-Unconstrained scenarios, showing severe performance degradations whenever their input distribution changes. 
We then show how single source DG algorithms coupled with OWR methods can mitigate the domain-shift problem, despite being still far from solving it (Section \ref{exp:owr-dg}). We finally discuss the implications of our benchmark, open issues and future research directions (Section \ref{exp:owr-discussion}).

\subsection{Are OWR models Robust to Domain Shift?}
\label{exp:owr-alone}

\begin{figure*}[t]
    \vspace{5pt}
    \centering
    \begin{subfigure}{0.3\linewidth}
        \includegraphics[width=\linewidth]{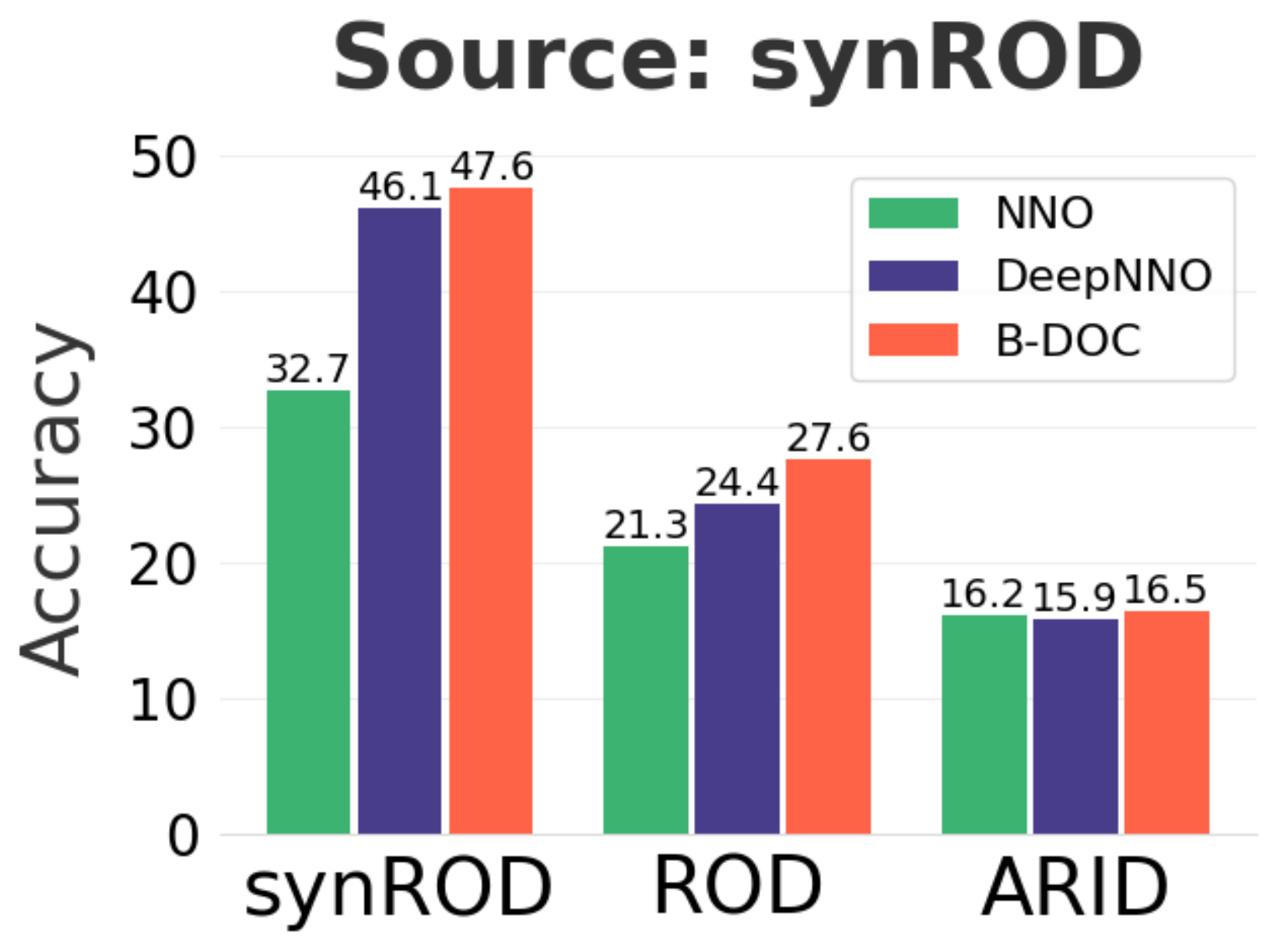}
        \caption{\centering{Closed World Without Rejection}\vspace{-5pt}}
        \label{fig:synROD-wor}
    \end{subfigure}
    \hfill
    \begin{subfigure}{0.3\linewidth}
        \includegraphics[width=\linewidth]{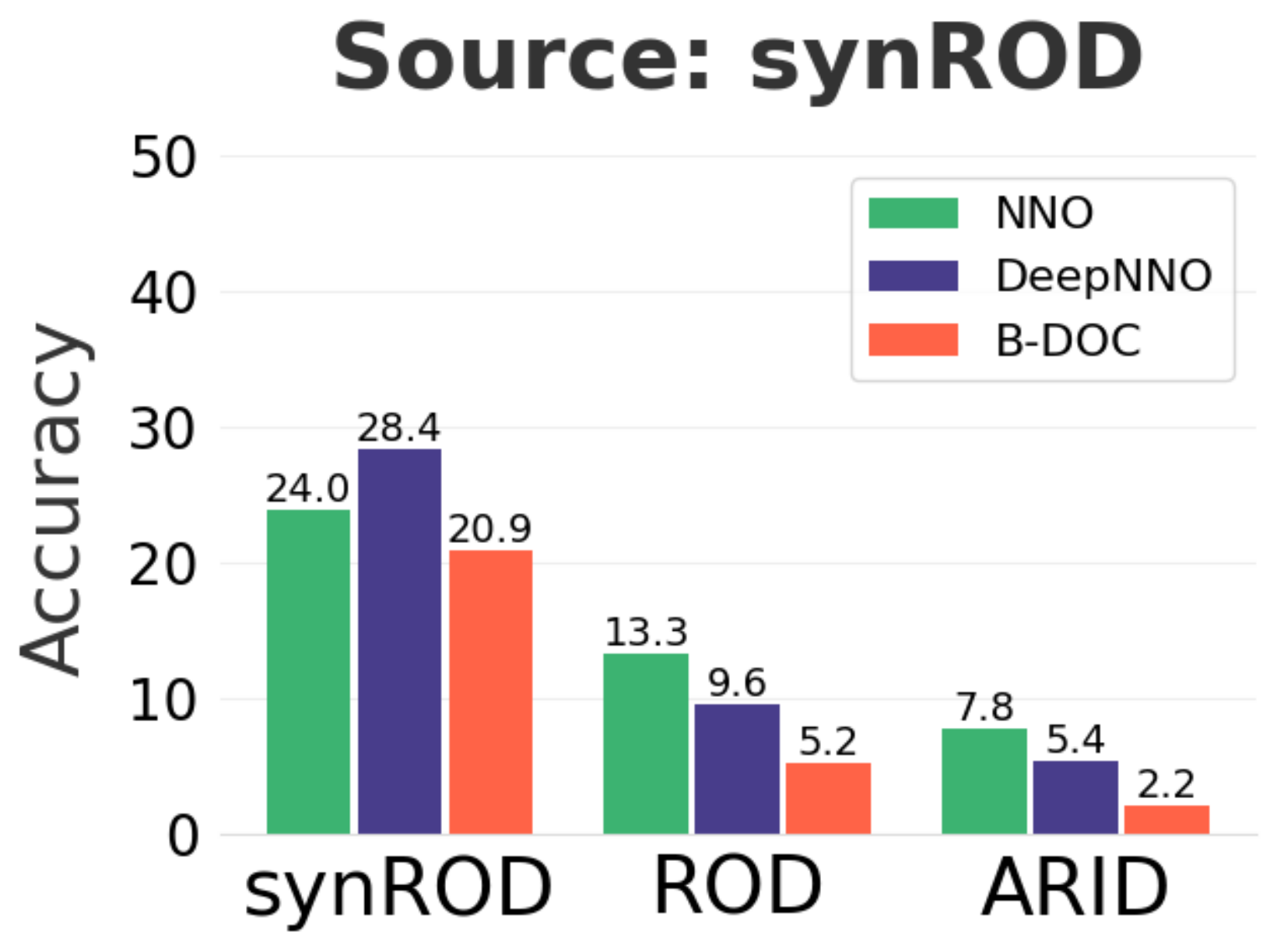}
        \caption{\centering{Closed World With Rejection}\vspace{1em}\vspace{-5pt}}
        \label{fig:synROD-wir}
    \end{subfigure}
    \hfill
        \begin{subfigure}{0.3\linewidth}
        \includegraphics[width=\linewidth]{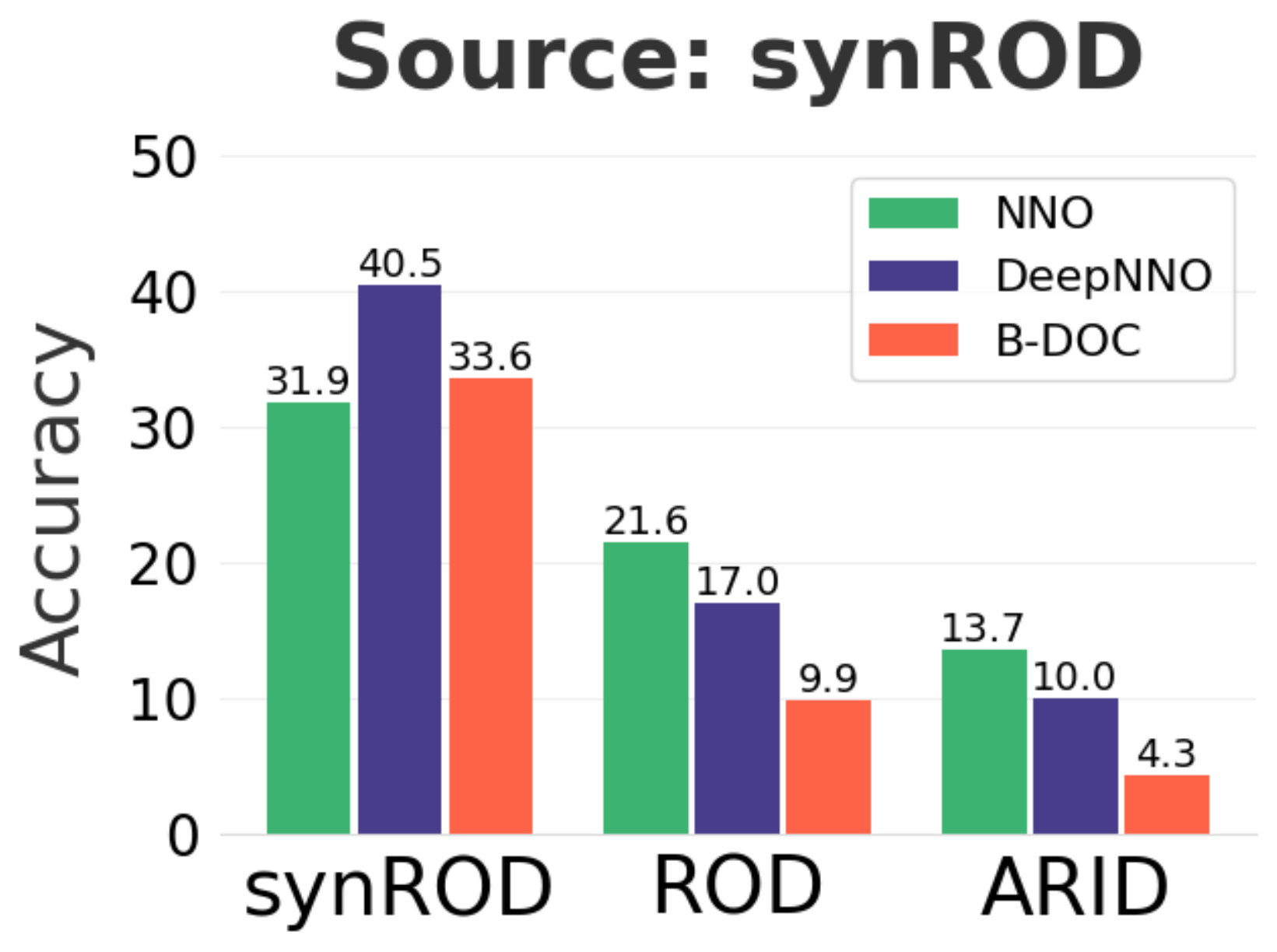}
        \caption{\centering{OWR Harmonic mean}\vspace{1em}\vspace{-5pt}}
        \label{fig:synROD-h}
    \end{subfigure}
    \hfill
     \caption{Comparison of NNO \cite{bendale2015towards}, DeepNNO \cite{mancini2019knowledge} and B-DOC \cite{fontanel2020boosting} trained on synROD \cite{loghmani2020unsupervised} and tested on synROD \cite{loghmani2020unsupervised}, ROD \cite{lai2011large} and ARID \cite{loghmani2018recognizing}. The numbers denote the average accuracy among the different incremental steps. \vspace{-13pt}}
     \label{fig:synROD}
\end{figure*}

\begin{figure*}[t]
    \vspace{5pt}
    \centering
    \begin{subfigure}{0.3\linewidth}
        \centering
        \includegraphics[width=\linewidth]{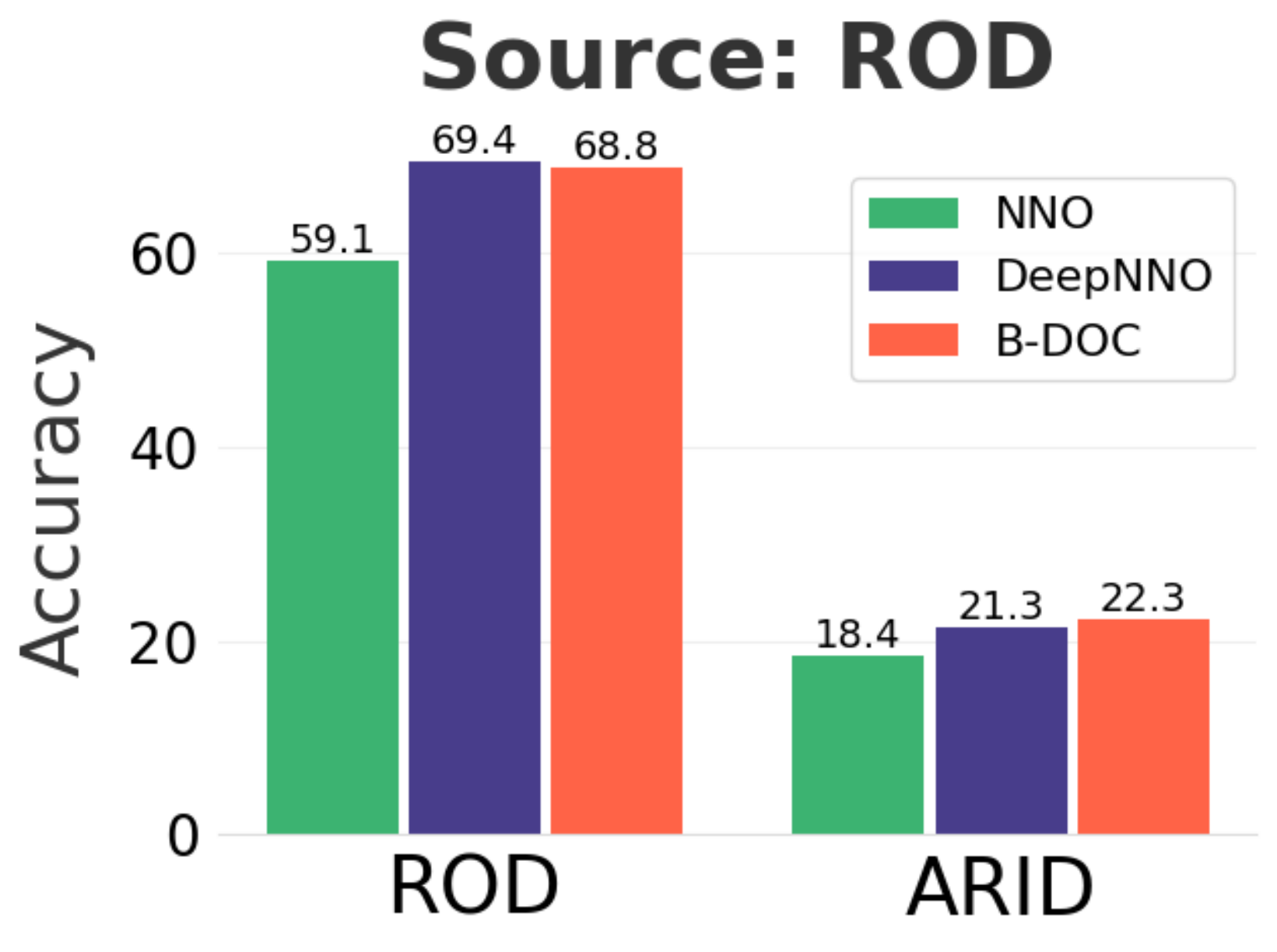}
        \caption{\centering{Closed World Without Rejection}\vspace{-5pt}}
        \label{fig:ROD-wor}
    \end{subfigure}
    \hfill
    \begin{subfigure}{0.3\linewidth}
        \centering
        \includegraphics[width=\linewidth]{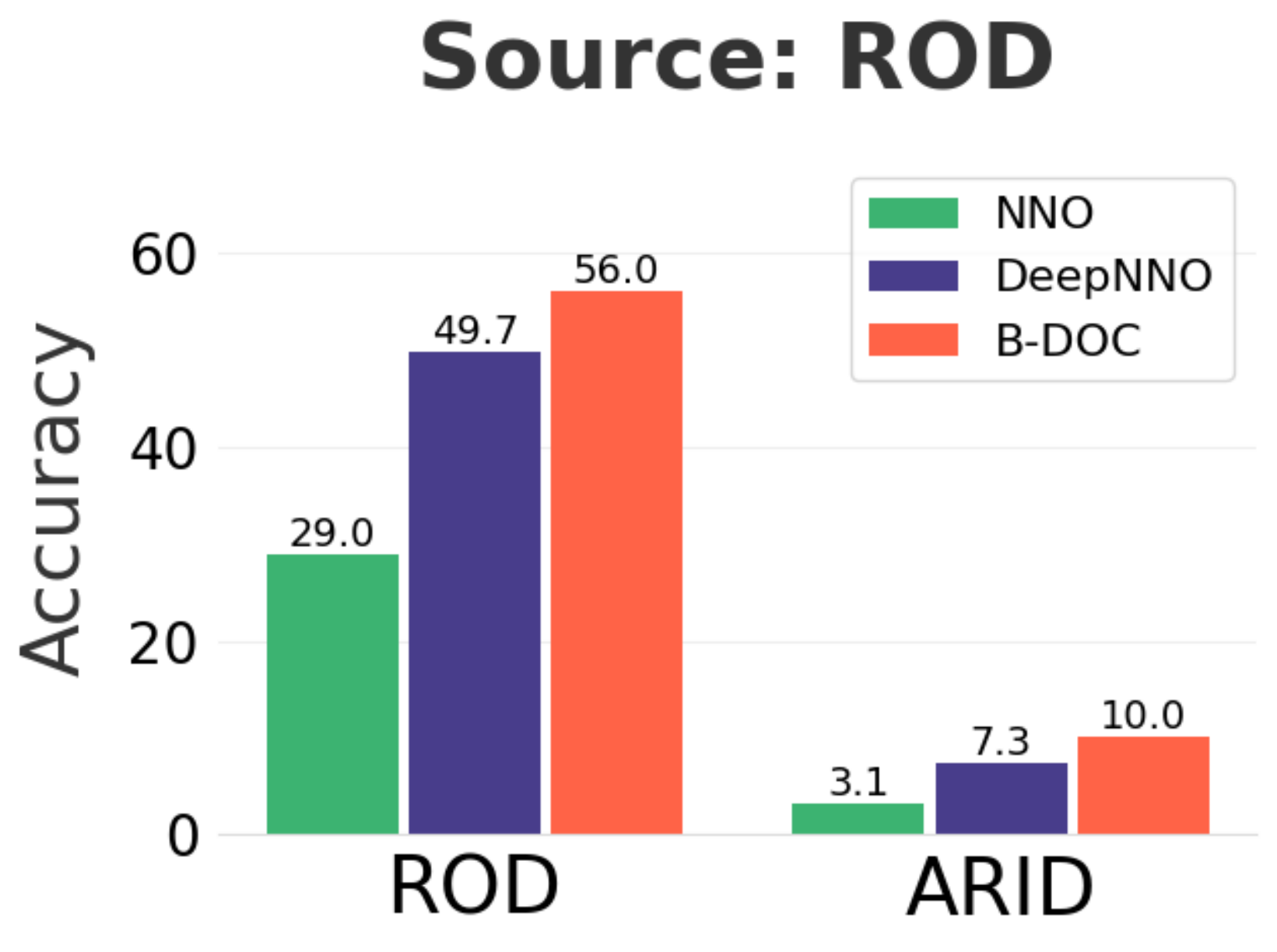}
        \caption{\centering{Closed World With Rejection}\vspace{1em}\vspace{-5pt}}
        \label{fig:ROD-wir}    
    \end{subfigure}
    \hfill
      \begin{subfigure}{0.3\linewidth}
        \centering
        \includegraphics[width=\linewidth]{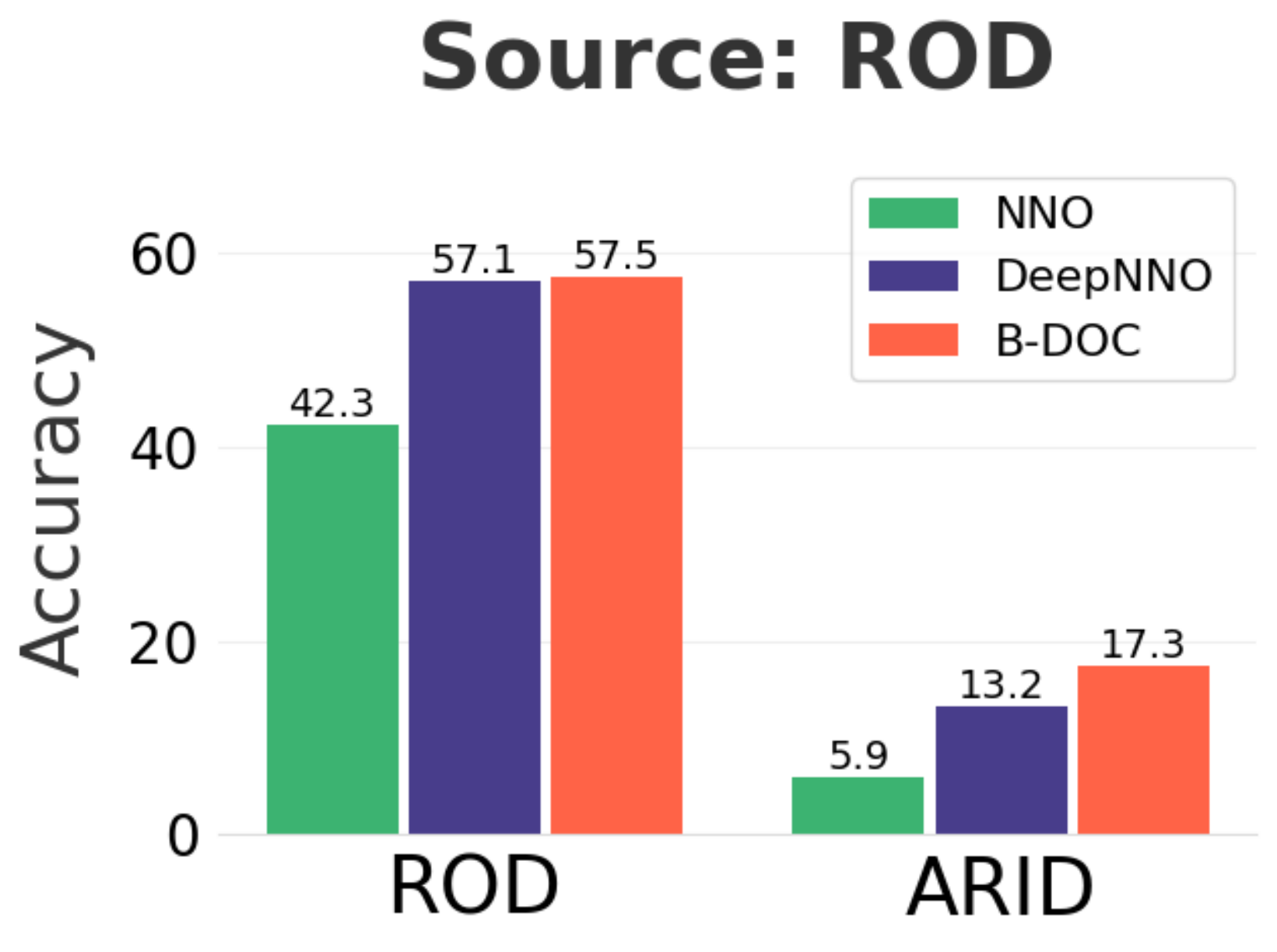}
        \caption{\centering{OWR Harmonic mean}\vspace{1em}\vspace{-5pt}}
        \label{fig:ROD-h}    
    \end{subfigure}
    \hfill
     \caption{Comparison of NNO \cite{bendale2015towards}, DeepNNO \cite{mancini2019knowledge} and B-DOC \cite{fontanel2020boosting} trained on ROD \cite{lai2011large} and tested on ROD \cite{lai2011large} and ARID \cite{loghmani2018recognizing}. The numbers denote the average accuracy among the different incremental steps. 
     \vspace{-13pt}}
    \label{fig:ROD}
 \end{figure*}
 
\myparagraph{Synthetic-to-Real.} We start our experimental analysis by considering as source domain the synROD dataset and all the others as target domains in turn. Results are reported in Fig.~\ref{fig:synROD} in terms of closed world without rejection performances, closed world with rejection and OWR harmonic mean. 
As we can see from Fig.~\ref{fig:synROD}, all the OWR methods suffer a significant drop in performance under domain-shift. In particular, from Fig.~\ref{fig:synROD-wor} we note that in the closed world scenario \textit{without} the possibility of classifying samples as unknowns, recognizing real objects is a very difficult task for all the OWR methods trained on synthetic data. While DeepNNO and B-DOC achieve good results without domain-shift, (47.6\% and 46.1\% respectively) their performance drop of almost 18\% when going from synROD to ROD, and of almost 26\% from synROD to the more challenging ARID. 

Similarly, in Fig.~\ref{fig:synROD-wir}, with the rejection option the performances drop in average of almost 15\% on ROD and of more than 19\% on ARID. Surprisingly, 
B-DOC  suffers more than all the others when the rejection option is introduced, losing nearly 16\% accuracy on ROD and 20\% on ARID. 
This may be due to the fact that the thresholds of B-DOC are estimated on a held out set from the training data. Consequently, since these samples do not represent the distribution of test samples under domain-shifts, the wrongly computed thresholds lead the model to perform poorly. Similar is the behaviour of the other deep model, DeepNNO, reaching the poor closed world with rejection accuracies of 9.6\% on ROD and 5.4\% on ARID. Surprisingly, the non end-to-end approach NNO  achieves considerably higher performances on ROD and ARID, with an average loss of 13.5\%. 
The reason behind this behaviour is that the threshold computed by NNO on synROD is usually low, due to the low confidence on the predictions of its shallow classification model and the high variability of the dataset. This allows NNO to reject less samples, thus better preserving the closed world accuracy. Despite that, the performance on ARID, with an average of nearly 5\%, arise serious concerns on the applicability of these algorithms in real scenarios. 

Finally, as a global analysis from Fig. \ref{fig:synROD-h}, we can see that the OWR-H performances confirm the previous trends. All the methods suffer a huge performance drop, of more than 19\% on average on ROD and of 26\% in ARID. In particular, the performance of both deep models, DeepNNO and B-DOC decrease of almost 23\% in ROD and of 30\% in ARID. Again (and surprisingly) NNO shows a good trade-off, with a decrease in performance of almost 15\% on average. 

\myparagraph{Constrained-to-Unconstrained.}  We continue our experimental analysis in a different, real-to-real scenario, by considering as source domain ROD and as target domains ROD itself and ARID. We highlight that both ROD and ARID are real datasets and their shift being only on the environments they depict, constrained ROD and unconstrained ARID. 
 Fig.~\ref{fig:ROD} shows the results. As expected, while they achieve good performance on the same domain, all the methods suffer a considerable drop in performances when tested under domain shifts. However, this drop is even larger than the one experienced in the synthetic-to-real case. In the closed world with rejection (Fig.~\ref{fig:ROD-wor}), the models which had an average accuracy of 65\% on ROD, lose almost 45\% of accuracy when tested on ARID, obtaining performances close to 20\% accuracy. The same drop in accuracy happens in the closed world with rejection scenario (Fig.~\ref{fig:ROD-wir}). The domain-shift leads the models to confuse samples of the new domains as unknowns, as demonstrated by the drop in performance between ROD and ARID: in the latter, the accuracy with rejection is barely 7\% on average.

The overall performances with the OWR-H metric are very unsatisfactory and there is a significant gap (almost 40\% on average) between the accuracy values reached on ROD and ARID, considering all the three methods. These results highlight how they significantly suffer when tested on data belonging to new domains/environments and confirms our conjecture that domain-shift is a huge problem for OWR algorithms. 

\begin{figure*}[t]
\vspace{5pt}
    \centering
    \begin{subfigure}{0.3\linewidth}
        \centering
        \includegraphics[width=\linewidth]{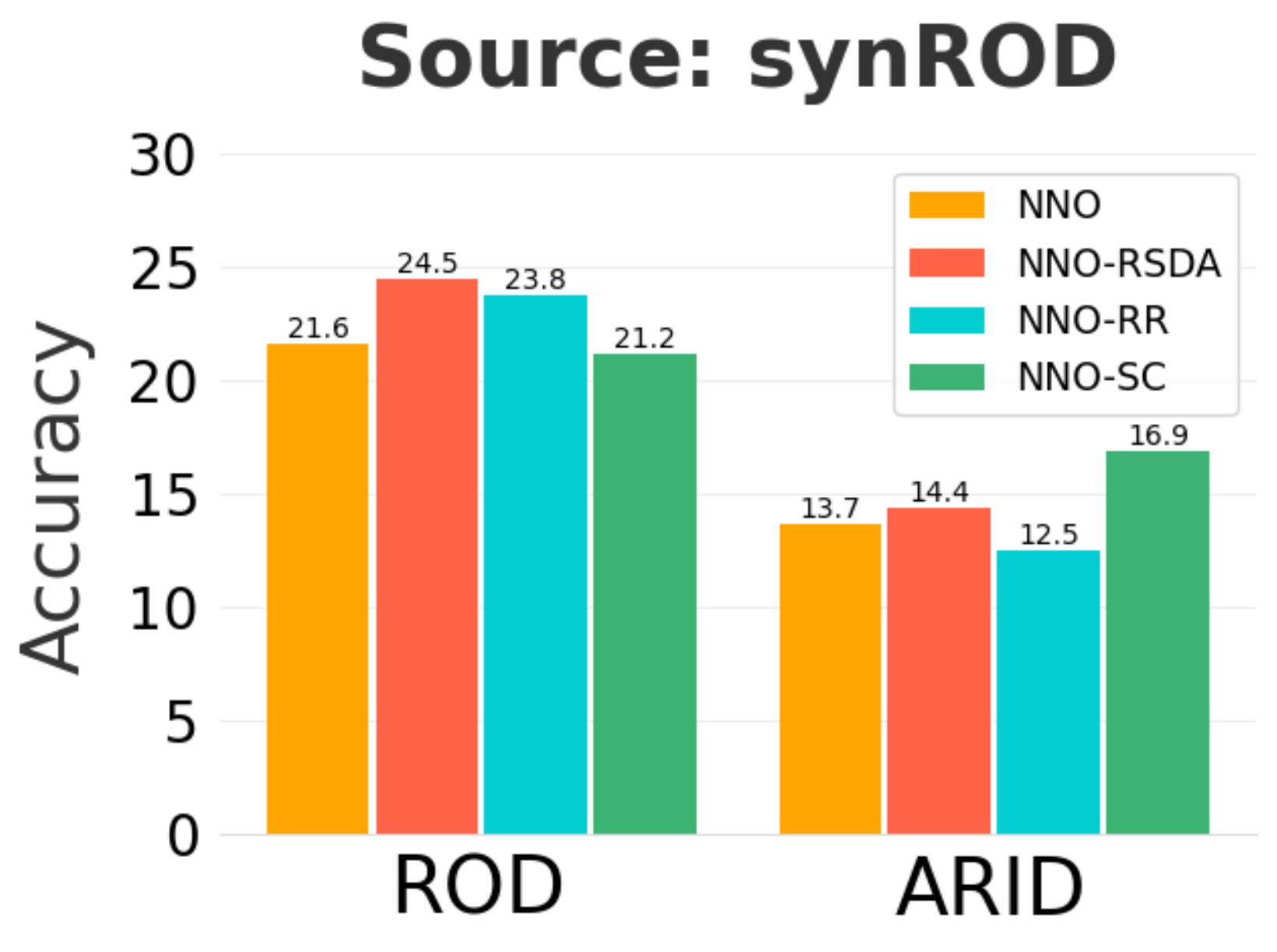}
        \caption{\centering{NNO - OWR Harmonic}\vspace{-5pt}}
        \label{fig:synrod-nno-dg}
    \end{subfigure}
    \hfill
    \begin{subfigure}{0.3\linewidth}
        \centering
        \includegraphics[width=\linewidth]{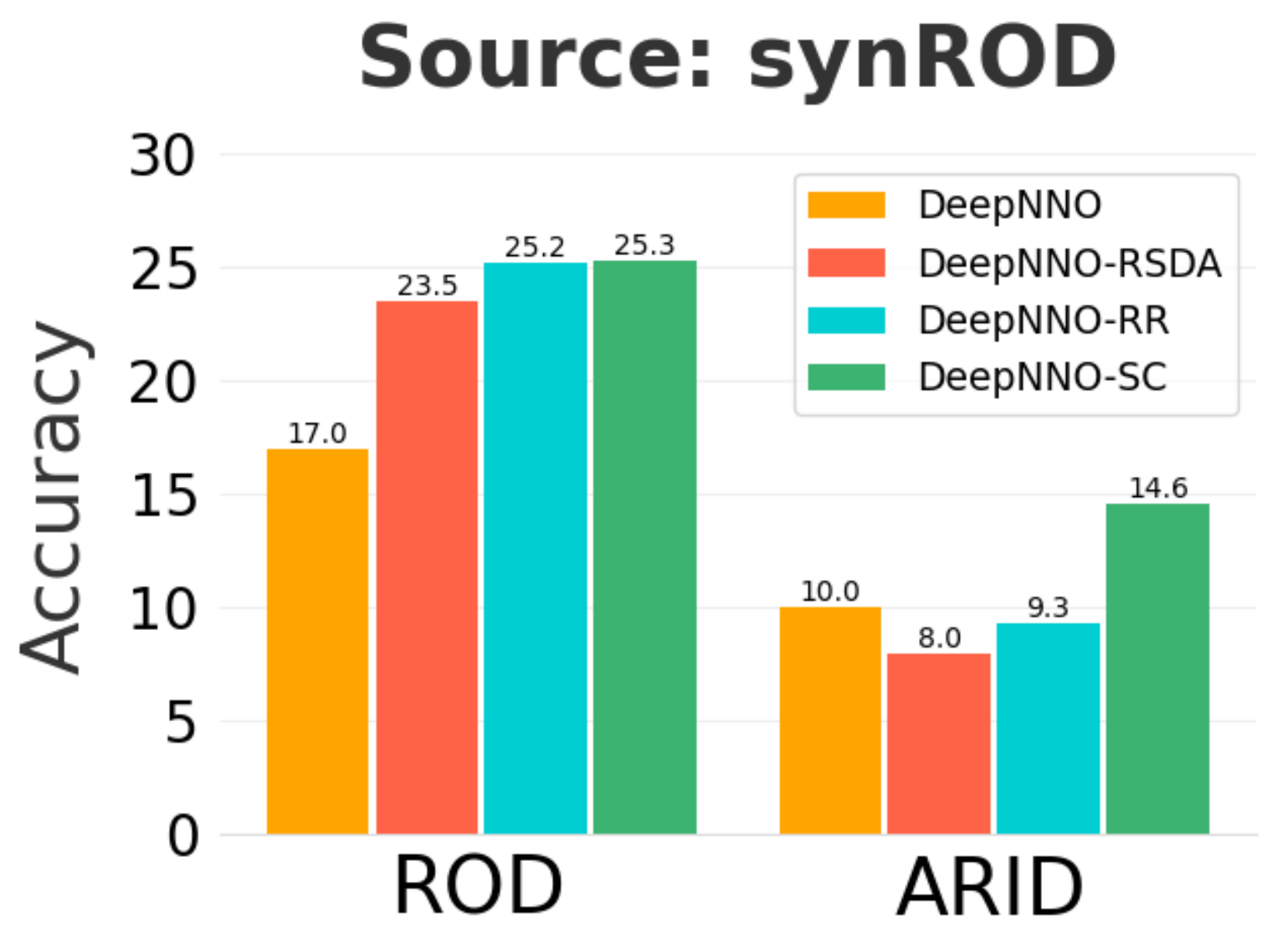}
        \caption{\centering{DeepNNO - OWR Harmonic}\vspace{-5pt}}
        \label{fig:synrod-deepnno-dg}
    \end{subfigure}
    \hfill
        \begin{subfigure}{0.3\linewidth}
        \centering
        \includegraphics[width=\linewidth]{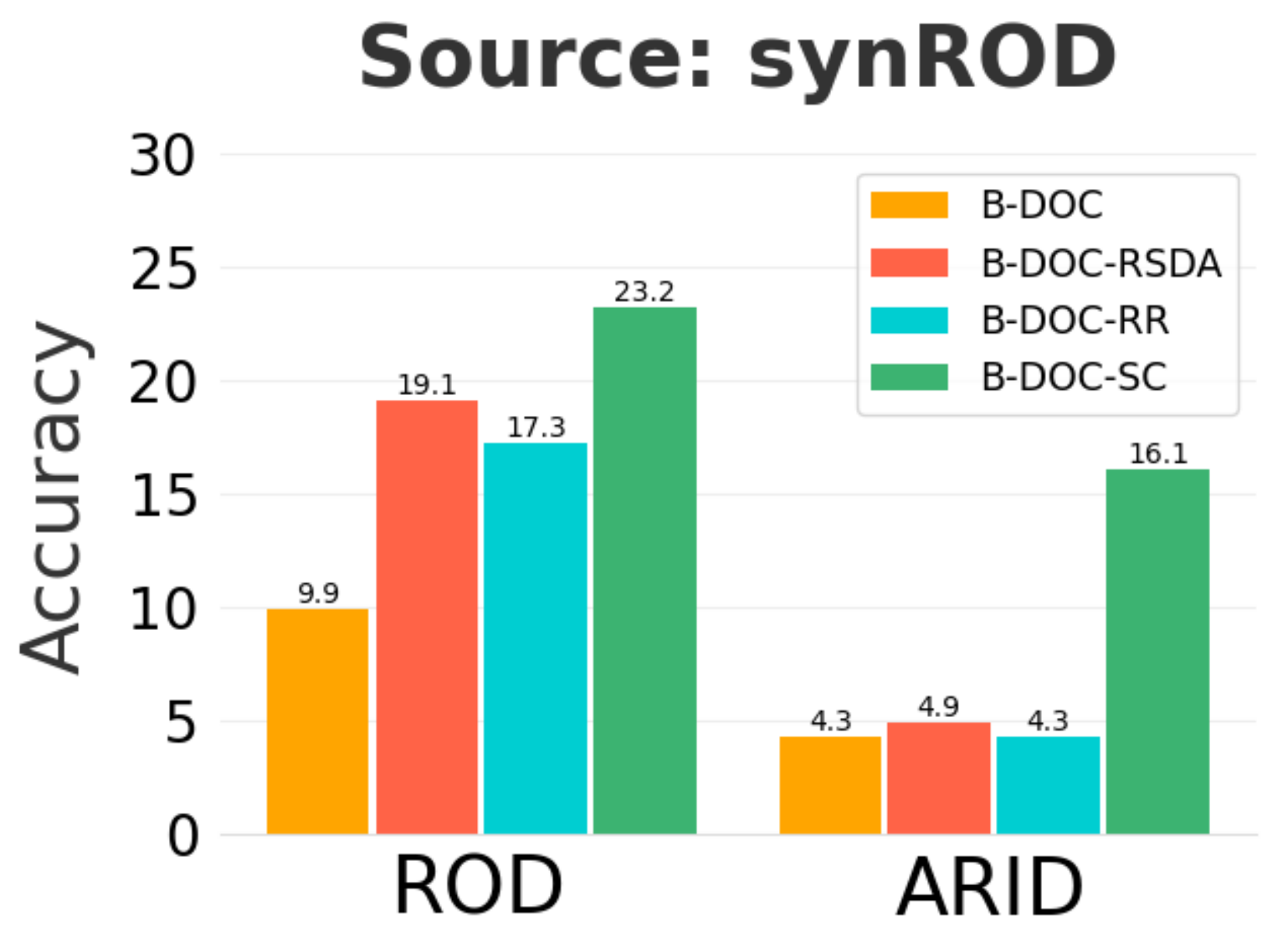}
        \caption{\centering{B-DOC - OWR Harmonic}\vspace{-5pt}}
        \label{fig:synrod-bdoc-dg}
    \end{subfigure}
    \hfill
     \caption{Comparison of NNO \cite{bendale2015towards}, DeepNNO \cite{mancini2019knowledge} and B-DOC \cite{fontanel2020boosting} with Domain Generalization techniques when trained on synROD \cite{loghmani2020unsupervised} and tested on ROD \cite{lai2011large} and ARID \cite{loghmani2018recognizing}. The numbers denote the average accuracy among the different incremental steps. \vspace{-13pt}}
     \label{fig:synROD-DG}
\end{figure*}

\begin{figure*}[t]
\vspace{5pt}
    \centering
     \begin{subfigure}{0.3\linewidth}
        \centering
        \includegraphics[width=\linewidth]{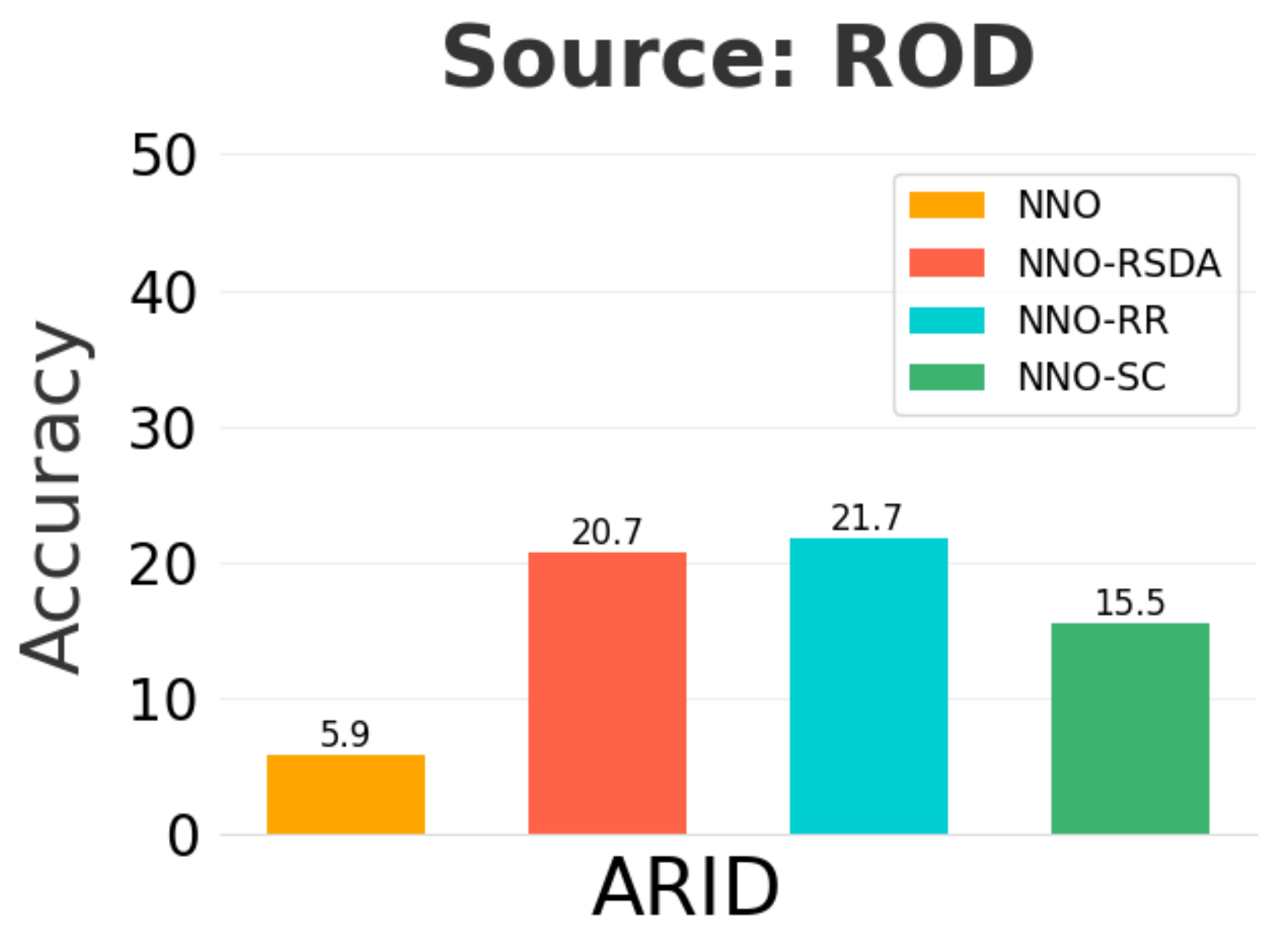}
        \caption{\centering{NNO - OWR Harmonic}\vspace{-5pt}}
        \label{fig:rod-nno-dg}
    \end{subfigure}
    \hfill
    \begin{subfigure}{0.3\linewidth}
        \centering
        \includegraphics[width=\linewidth]{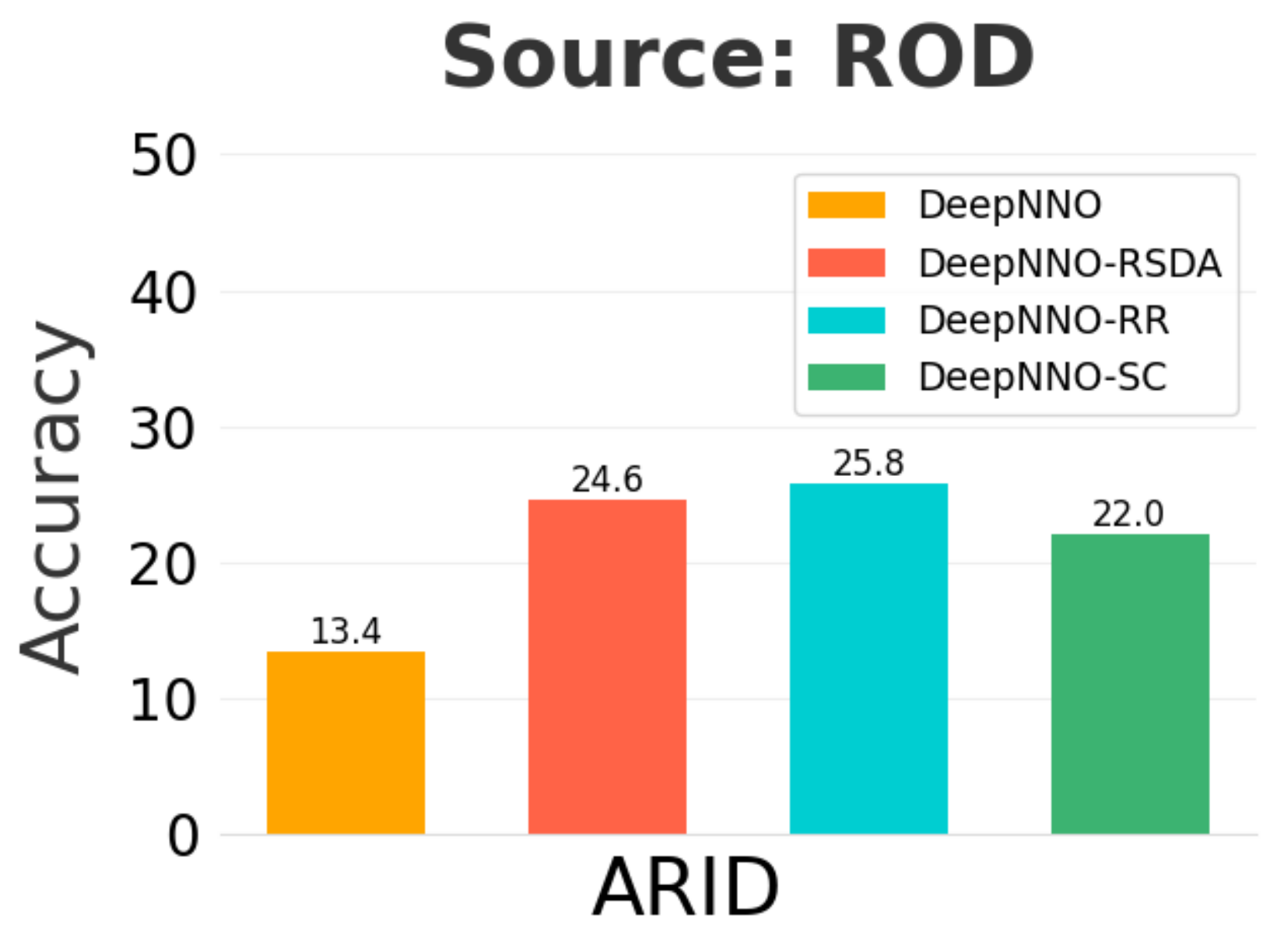}
        \caption{\centering{DeepNNO - OWR Harmonic}\vspace{-5pt}}
        \label{fig:rod-deepnno-dg}
    \end{subfigure}
    \hfill
    \begin{subfigure}{0.3\linewidth}
        \centering
        \includegraphics[width=\linewidth]{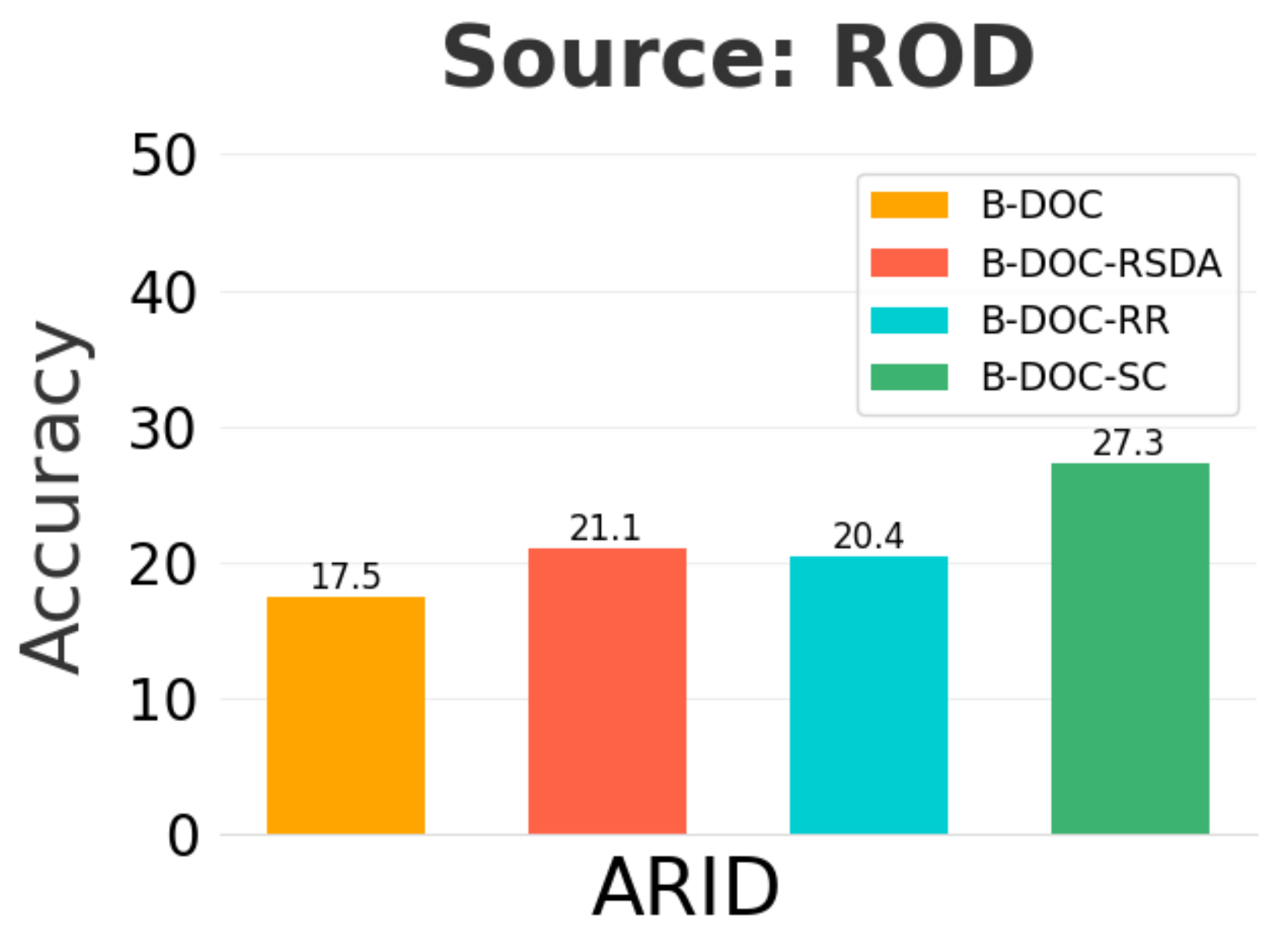}
        \caption{\centering{B-DOC - OWR Harmonic}\vspace{-5pt}}
        \label{fig:rod-bdoc-dg}
    \end{subfigure}
    \hfill
     \caption{Comparison of NNO \cite{bendale2015towards}, DeepNNO \cite{mancini2019knowledge} and B-DOC \cite{fontanel2020boosting} with Domain Generalization techniques when trained on ROD \cite{loghmani2020unsupervised} and tested on ARID \cite{loghmani2018recognizing}. The numbers denote the average accuracy among the different incremental steps. \vspace{-13pt}}
     \label{fig:ROD-DG}
\end{figure*}

\subsection{Can DG methods address the problem?}
\label{exp:owr-dg}
Given the poor results of OWR methods under domain-shift, in this section we check whether single source DG algorithms can be used to address this problem. For sake of space, we consider the OWR-H metric for the comparisons.

\myparagraph{Synthetic-to-Real.} We start with the synthetic-to-real scenario. Fig.~\ref{fig:synROD-DG} shows how the OWR-H of all methods changes when equipped with a DG algorithm. 
While NNO experiences a slight improvement, the performances of DeepNNO and B-DOC largely benefit from DG techniques, especially when the SC strategy \cite{huang2020selfchallenging} is applied. Indeed, the results on ARID are 2 and 4 times higher than the originals for DeepNNO and B-DOC respectively. Looking at the other DG methods, they all improve DeepNNO and B-DOC on ROD, while their performance on ARID varies, bringing no to little gains. We ascribe the higher effectiveness of SC to the fact that it regularizes the classifier, forcing it to i) focus on multiple cues and ii) achieve a lower confidence on the predictions which might help in estimating better thresholds for detecting unknown samples in different domains.  
As stated above, for NNO, all the DG techniques have low to even negative effect (Fig.~\ref{fig:synrod-nno-dg}). This happens because NNO does not fine-tune its representation across multiple training stages, thus the DG method has a more limited impact. 

For what concerns the other DG methods, while RR  \cite{bucci2020rotation} brings a slight improvement on almost all baselines regardless of the scenario, RSDA \cite{volpi2019addressing} is more effective on ROD rather then ARID. The reason is that synROD images differ from ROD ones mainly in colors and shapes, thus using precise data augmentation to bridge these differences leads to a general improvement in performance. For ARID, this occurs only partially due to other (different challenges) such as occlusion and scale variations. 

Despite these results, the domain-shift problem is still heavily present, with an average drop from the synROD performance of almost 10\% on ROD and of 19\% on ARID. 

\myparagraph{Constrained-to-Unconstrained.} Finally, in this section we analyze the effectiveness of DG techniques for models trained on ROD and tested on ARID. 
We report in Fig. \ref{fig:ROD-DG} the results. As the figure shows, all the DG methods bring improvements to OWR methods in this scenario, even if in different ways. For instance, SC leads to the highest performances only for B-DOC, with 27.3\% of accuracy, while for DeepNNO and NNO, RSDA and RR are more effective. In particular, despite the limited training phase in which the DG algorithms act, NNO still benefits from them in this scenario, with performances more than 3 times higher when coupled with RR and RSDA. This means that the augmentations these methods apply are effective in reducing the domain-shift between ROD and ARID. However, compared to the original results on ROD, the gaps are still impressive: NNO+RR is still 40\% lower than NNO applied on ROD. Similarly, DeepNNO+RR and B-DOC+SC are almost 30\% far from their counterpart applied on ROD. These results confirm that the domain-shift in OWR can be mitigated (but not solved) using single-source DG algorithms. 

\subsection{Discussion and future directions}
\label{exp:owr-discussion} 




From the results of Section~\ref{sec:experiments}, we can draw two important conclusions. First, OWR methods are not robust to domain-shift, performing much worse when tested on input distributions different from the training ones. Second, despite the slight improvements, coupling methods for single-source DG with OWR techniques is not sufficient to solve this problem. After these conclusions, we want to highlight two open issues which we believe the community should focus on to produce OWR systems applicable in the real-world.

\myparagraph{Domain-shift in recognition.} OWR models need to be capable of adapting to unseen domains quickly when applied in the wild. In this work we highlighted how a combination of DG and OWR methods is not enough to tackle this issue, with poor recognition performances and wrong estimation of the rejection thresholds. Future works could develop a single model for the two problems and/or couple them with algorithms exploiting the incoming stream of target data, in an online DA fashion \cite{mancini2018kitting}.

\myparagraph{Domain-shift while learning.} Current OWR algorithms assume that data in each incremental step arrive from the same input distribution. An open question is what would happen if different incremental steps contain data from different domains. Our intuition is that, while being more prone to forgetting, a model will also tend to use domain cues to perform a (wrong) prediction, which is something we definitely want to avoid. Disentangling domain-specific and semantic-specific information in this scenario is very challenging and it might require either the use of unlabeled data \cite{kundu2020class} or side-information \cite{mancini2020dgzsl}.

\section{Conclusion and Future Works} 
\label{sec:conlusions} 
In this work we tested the robustness of OWR algorithms to domain-shift. In particular, we proposed a cross-domain benchmark for OWR containing multiple classes and acquisition conditions. We then tested OWR algorithms in this benchmark showing how: i) they heavily suffer from the domain-shift problem; and ii) coupling them with DG algorithms only mitigates this issue, being far from solving it. According to the results, we highlighted some open issues and future directions in this topic.

Finally, while these are some open issues of OWR systems, there are others not strictly related to domain-shift that remain to be solved. An example is how to make the model autonomously learn from the environment, inferring the label of a detected unknown object and collecting the data necessary to learn it, even from the web \cite{mancini2019knowledge}.  We believe our results and considerations show how a collective effort is needed to have practical and autonomous OWR systems able to reliably act in the real world.

\bibliography{bib}

\end{document}